\documentclass[twoside]{article}
\usepackage[accepted]{aistats2015}
\usepackage{amsfonts}
\usepackage{nicefrac}
\usepackage{xspace}
\usepackage{graphicx}
\graphicspath{{img/}}
\usepackage{multirow}
\usepackage{macros}
\usepackage{url}
\usepackage{algorithm}
\usepackage{algorithmicx}
\usepackage{algpseudocode}
\usepackage{natbib}
\usepackage{bm}
\bibpunct{(}{)}{;}{a}{,}{,}
\usepackage{color}
\begin{document}

%

%

\twocolumn[

\aistatstitle{DART: Dropouts meet Multiple Additive Regression Trees}

\aistatsauthor{ K. V. Rashmi \And Ran Gilad-Bachrach}

\aistatsaddress{Department of Electrical Engineering and Computer Science~~ \\UC Berkeley \And Machine Learning Department\\Microsoft Research} ]
\begin{abstract}
MART \citep{friedman2001greedy,friedman2002stochastic}, an ensemble model of boosted regression trees, is known to deliver high prediction accuracy for diverse tasks, and it is widely used in practice. However, it suffers  an issue which we call \textit{\overspecialization}, wherein trees added at later iterations tend to impact the prediction of only a few instances, and make negligible contribution towards the remaining instances. This negatively affects the performance of the model on unseen data, and also makes the model over-sensitive to the contributions of the few, initially added tress. We show that the commonly used tool to address this issue, that of \textit{shrinkage}, alleviates the problem only to a certain extent and the fundamental issue of \overspecialization still remains. 

In this work, we explore a different approach to address the problem that of employing \textit{dropouts}, a tool that has been recently proposed in the context of learning deep neural networks \citep{hinton2012improving}. We propose a novel way of employing dropouts in MART, resulting in the \textit{\algname} algorithm. We evaluate \algname on ranking, regression and classification tasks, using large scale, publicly available datasets, and show that \algname outperforms MART in each of the tasks, with a significant margin. We also show that \algname overcomes the issue of \overspecialization to a considerable extent.

\end{abstract}

\section{Introduction} \label{sec:intro}
Ensemble based algorithms have been shown to achieve high accuracy for a number of machine learning tasks \citep{caruana2006empirical}. For ensembles to achieve better accuracy than the individual predictors that they are made of, these predictors need to be accurate but uncorrelated \citep{breiman2001random}.
This helps to increase the accuracy of the model by reducing the sensitivity to specific features or instances that might exist in the individual predictors~\citep{breiman2001random,hinton2012improving}. While some classes of ensemble algorithms such as random forests~\citep{breiman2001random} learn each predictor in the ensemble independently, boosted ensemble algorithms such as AdaBoost \citep{freund1995desicion} and MART \citep{friedman2001greedy,friedman2002stochastic}\footnote{This algorithm is known by many names, including Gradient TreeBoost, boosted trees, and Multiple Additive Regression Trees (MART). We use the latter to refer to this algorithm.} iteratively add each predictor. 

Boosting algorithms add predictors that focus on improving the current model, and this is achieved by modifying the learning problem between iterations. While this guarantees that the added predictor is different than the ones in the ensemble, the new predictors typically focus on a small subset of the problem and hence do not have a strong predictive power when measured on the original problem. This increases the risk of adding models that over-fit specific instances. This is a well-known problem in the context of boosting~\citep{freund2001adaptive} as well as in MART, which is an ensemble of boosted regression trees. Here, trees added at later iterations tend to impact the prediction of only a few instances, and they make negligible contribution towards the prediction of all the remaining instances. This, in turn, can negatively impact the performance of the algorithm on unseen data by increasing the capacity of the model without making significant improvement in its training error. This also makes the model over-sensitive to the contributions of the few, initially added tress. We call this issue of subsequent trees affecting the prediction of only a small fraction of the training instances \textit{\overspecialization}. We discuss this issue in greater detail in Section~\ref{sec:overfitting} with an example from a regression task on a real-world dataset.  

The most common approach employed to combat the problem of \overspecialization in MART is  \textit{shrinkage}~\citep{friedman2001greedy,friedman2002stochastic}. Here, the contribution of each new tree is reduced by a constant value called the \emph{shirnkage factor}.  As we will see in Section~\ref{sec:overfitting}, shrinkage does help in
reducing the impact of the first trees, nevertheless, however, as the size of the ensemble increases, the problem of \overspecialization reappears. 

In this work, we explore a different approach to address the issue of \overspecialization in MART. We propose employing \textit{dropouts}, a tool that has been recently proposed in the context of learning deep neural networks \citep{hinton2012improving}. 
In neural networks, dropouts are used to mute a random fraction of the neural connections during the learning process. Therefore, nodes at higher layers of the network cannot rely on a few connections to deliver the information needed for the prediction. This method has contributed significantly to the success of deep neural networks for many tasks including, for example,  
object classification in images \citep{krizhevsky2012imagenet}. 

The technique of dropouts has been used successfully in other learning models \citep{ICML2013_vandermaaten13, wang2013fast}, for example, in logistic regression \citep{wager2013dropout}. In these cases, dropouts are used to mute a random fraction of the input features during the training phase. In the context of ensemble of trees, this approach makes them similar to the approach employed by random forests for diversification~\citep{breiman2001random}, wherein each tree in the ensemble is learned (independently) using a different random fraction of the features. 

In this paper, we propose a novel way of employing dropouts for ensemble of trees: muting complete trees as opposed to muting features.\footnote{Muting trees and muting features can be done at the same time and indeed we do this in our experiments.} We employ this approach in MART and call the resulting algorithm \textit{\algname}. We evaluate \algname on three different tasks: ranking, regression and classification, using large scale, publicly available datasets. Our results show that \algname outperforms MART and random forest in each of the tasks, with significant margins (see Section~\ref{sec:eval}). We note that both MART and random forest are  known to be  highly successful models for many learning tasks \citep{caruana2006empirical}, for example, the winners of the `Yahoo! learning to  rank' challenge employed a MART model \citep{chapelle2011yahoo}. Therefore, it is both surprising and encouraging that we can squeeze out even higher accuracy out of MART. One of the reasons for the improved performance of \algname is that it addresses the issue of \overspecialization and results in more balanced contribution from all the trees in the ensemble (see Section~\ref{sec:overfitting}).

\section{Overcoming the \Overspecialization in MART} \label{sec:overfitting}
As we briefly discussed in Section~\ref{sec:intro}, boosting, in particular the MART algorithm, suffers from the issue of over-specialization: trees added at later iterations tend to impact the prediction of only a few instances, and make negligible contribution towards the prediction of all the remaining instances. In this section, we will demonstrate this issue and the impact of using dropouts as employed in \algname through an example from a regression task on the CTSlice data (see Section~\ref{sec:regression} for a description of the dataset and the task). We note that similar observations were made on the other datasets used in the evaluation (Section~\ref{sec:eval}) as well.

\begin{figure*}
\centering
\includegraphics[width=0.5\textwidth]{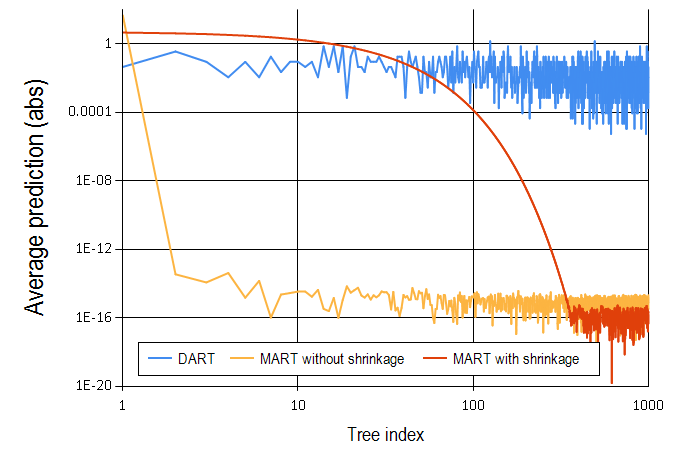}
\caption{The average contribution of the trees in the ensemble for different learning algorithms (the graph presents the absolute value of the average). The shrinkage factor used is 0.1. \label{img:contribution} }
\end{figure*}

\begin{figure*}
\centering
\begin{tabular}{|c c c c|}
\hline 
Index & MART without Shrinkage & MART with Shrinkage & DART \tabularnewline
\hline 
1 & \includegraphics[width=0.22\textwidth]{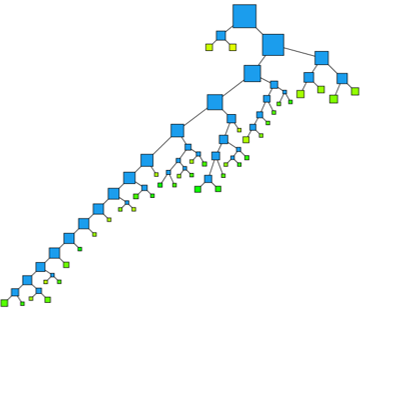} & \includegraphics[width=0.22\textwidth]{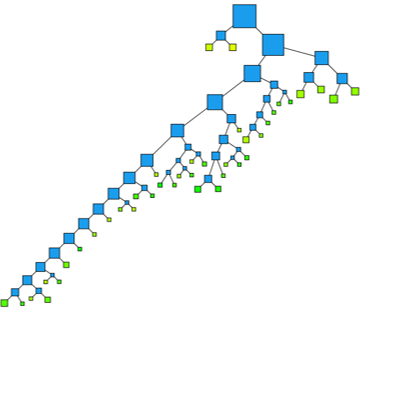} & \includegraphics[width=0.22\textwidth]{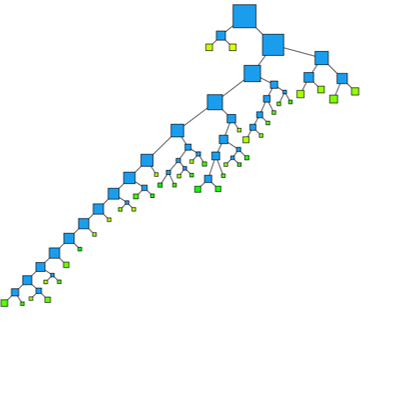}\tabularnewline
100 & \includegraphics[width=0.22\textwidth]{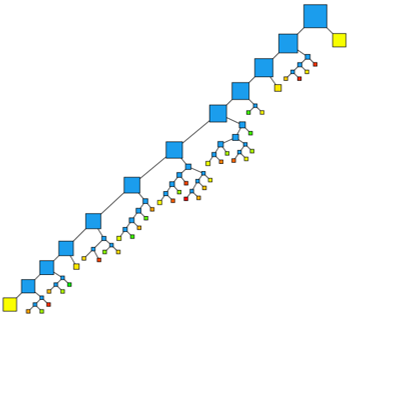} & \includegraphics[width=0.22\textwidth]{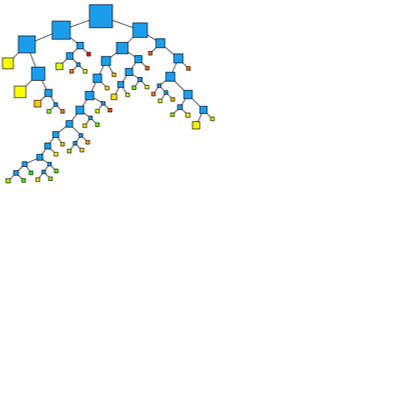} & \includegraphics[width=0.22\textwidth]{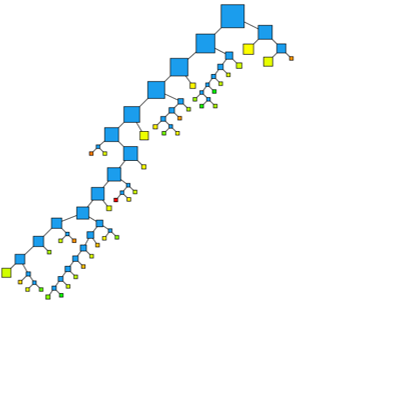}\tabularnewline
200 & \includegraphics[width=0.22\textwidth]{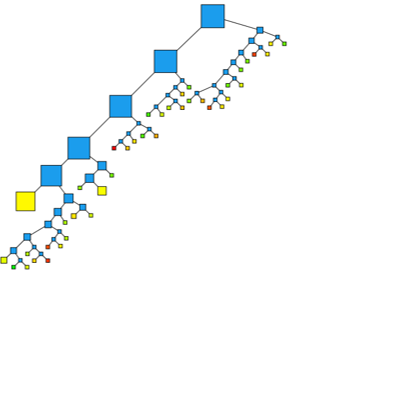} & \includegraphics[width=0.22\textwidth]{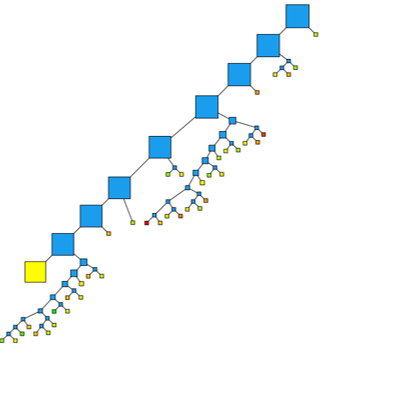} & \includegraphics[width=0.22\textwidth]{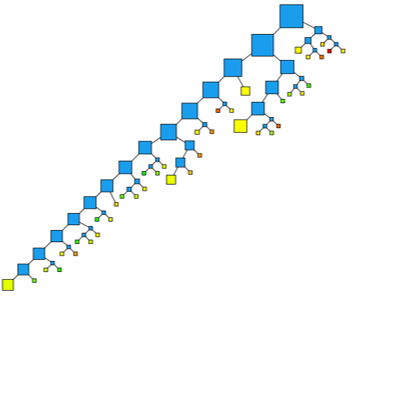}\tabularnewline
400 & \includegraphics[width=0.22\textwidth]{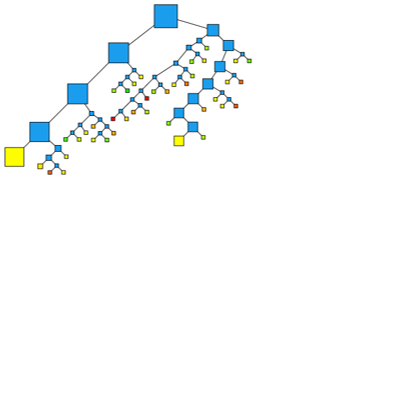} & \includegraphics[width=0.22\textwidth]{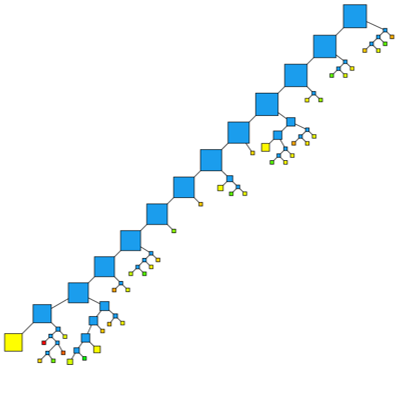} & \includegraphics[width=0.22\textwidth]{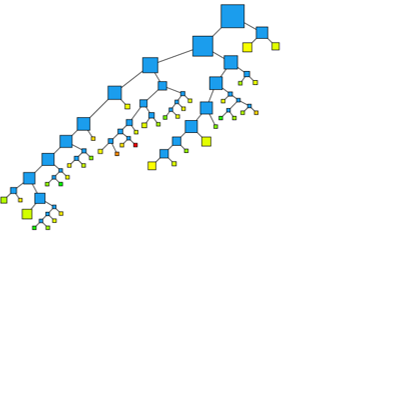}\tabularnewline
1000 & \includegraphics[width=0.22\textwidth]{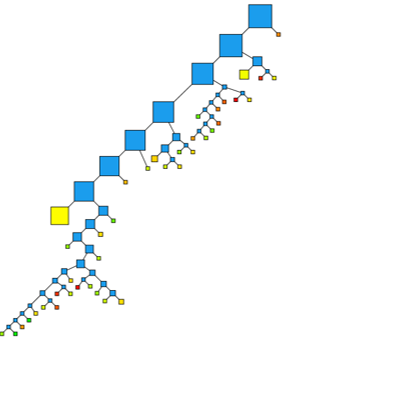} & \includegraphics[width=0.22\textwidth]{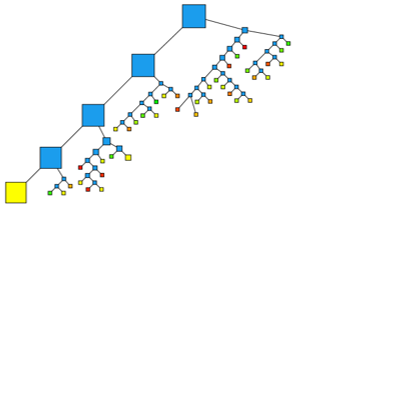} & \includegraphics[width=0.22\textwidth]{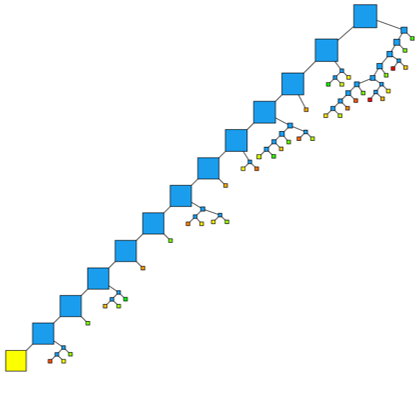}\tabularnewline
\hline 
\end{tabular}
\caption{Examples of trees in the ensemble for the regression task on CT slice dataset (Section~\ref{sec:regression}). Each column represents a different learning algorithm (MART (without shrinkage), MART+shrinkage, and DART). Each row represents a different index in the ensemble: 1st, 100th, 200th, 400th and 1000th tree in the ensemble. In each tree, the size of nodes is proportional to the percentage of the instances that reach this node. The color gradient of leaves represent the range of values where green stands for the positive extreme, yellow for zero, and red for negative extreme. \label{img:trees}}
\end{figure*}

Figure~\ref{img:contribution} presents the average contribution of the trees in the ensemble, where the average contribution of a tree $T$ is defined to be $\left|E_x\left[T(x)\right]\right|$ with the expectation taken with respect to the training data.
We can see that the MART algorithm (without using shrinkage) starts with a single tree that makes significant contribution and the rest of the trees add negligible contributions. We observed that even if we replace the term $\left|E_x\left[T(x)\right]\right|$ with $E_x\left[\left|T(x)\right|\right]$, the first tree has orders of magnitude larger contribution than the rest of the trees in the ensemble.
This behavior is inherent in the algorithm: if one would add a constant value to all the labels in the training data, only the first tree will get modified (with this constant value added to all its leaves) and the rest of the trees will remain with a small contribution to the model. Therefore, in a sense, the first tree learns the bias of the problem while the rest of the trees in the ensemble learn the deviation from this bias. This makes the ensemble very sensitive to the decisions made by the first tree. This can be seen in Figure~\ref{img:trees} as well, which depicts a few trees in the ensemble trained by different methods for the above mentioned task. We can see that the MART algorithm (without using shrinkage) adds trees that make negligible contribution to the overall prediction for most of the data points as indicated by the large yellow leaves in the first column.

As discussed briefly in Section~\ref{sec:intro}, shrinkage \citep{friedman2001greedy,friedman2002stochastic} is the most common approach employed to combat the issue of \overspecialization. Since shrinkage reduces the impact of each tree by a constant value, the first tree cannot compensate for the entire bias of the problem. We can see the impact of this strategy in Figure~\ref{img:contribution} as well as in Figure~\ref{img:trees}. We observe that the contribution of later trees do drop, but at a much slower rate than in the case where shrinkage is not used. For example, while the contribution of the 100th tree in MART without shrinkage is about 15 orders of magnitude smaller than the contribution of the first tree, this factor in MART with shrinkage drops to ``only" 4 orders of magnitude. In figure ~\ref{img:trees} we see that the large yellow leaves, representing the fact that a tree ``abstains" on many of the instances, appear later in the ensemble.  As we can see, the differences in the contributions from the trees in the ensemble are more gradual when shrinkage is used, nevertheless they are still notable.

Now, let us see the effect of using dropouts as employed in \algname. The last column in Figure~\ref{img:trees} depicts trees learned by the DART algorithm. First, compared to MART and MART with shrinkage, we see that trees specialized at a significantly slower rate as indicated by the much slower emergence of large yellow leaves.
This can be seen in Figure~\ref{img:contribution} as well, where we see that the expected contribution of the trees added in later iterations do not drop much.\footnote{Linear regression on this data suggests that there might be a slow decline in the average contribution of the tress at a rate of $0.0003$.}
Therefore, the sensitivity to the contribution of the individual trees is drastically reduced. At the same time, unlike random forest, \algname continues to learn trees to compensate for the deficiencies of the existing trees in the ensemble. It, however, does so in a controlled manner to strike a balance between diversity and over-specialization. We will see in Section~\ref{sec:alg} that both MART and random forest can be viewed as extreme cases of the \algname algorithm.

\section{Description of the \algname Algorithm} \label{sec:alg}
We start our presentation with the MART algorithm as the foundation on which DART builds.
MART can be viewed as a gradient descent algorithm \citep{friedman2001greedy}: at every iteration, MART computes the derivative of the loss function for the current predictions and adds a regression tree that fits the inverse of these derivatives to the ensemble. More formally, the input to the algorithm includes a set of points and their labels, $(x,y)$, where the points $x$ are in some space $\mathcal{X}$ and the labels $y$ are in a label space. The algorithm also takes as input a loss generating function which is tuned to the task at hand (for example, regression, classification, ranking, etc.). Using the loss generating function and the labels, the algorithm defines the loss for every point $x$, $L_x:\mathcal{Y} \mapsto \real$ where $\mathcal{Y}$ is the prediction space, typically the reals. For example, if the task is regression then the loss may be defined as $L_x(\hat{y})=(\hat{y}-y)^2$ where $y$ is the true label of $x$.

At every iteration, let the current model be denoted by $M:\mathcal{X}\mapsto\mathcal{Y}$ and $M(x)$ denotes the prediction of the current model for point $x$. Let $L^\prime_x\left(M(x)\right)$ be the derivative of the loss function at $M(x)$. MART creates an intermediate dataset in which a new label, $-L^\prime_x\left(M(x)\right)$, is associated with every point $x$ in the training data. A tree is trained to predict this inverse derivative and added to the ensemble as a step in the inverse direction of the derivative (in order to minimize the loss). 

The choice of the loss makes the MART algorithm applicable to a variety of learning tasks. As discussed earlier, the squared loss is used for regression tasks. The logistic loss function is used for classification tasks. Here, the loss function is defined to be $L_x(\hat{y}) = \left(1 + \exp\left(\lambda y \hat{y}\right)\right)^{-1}$ where $\lambda$ is a parameter. For  ranking tasks, the loss function would depend on the relative ordering of the points in the predicted ranking. In our evaluation (Section~\ref{sec:eval}) for ranking tasks, we use the definition of the LambdaMart  method~\citep{burges2010ranknet}. The main idea here is to directly define the gradient of the loss function:
$$ L^\prime_x\left(M(x)\right) := \sum_{x^\prime} \frac{s\left(x,x^\prime\right)\lambda}{1+\exp\left(\lambda\left(M(x)-M(x^\prime)\right)\right)}$$ where $\lambda$ is a parameter and $s\left(x,x^\prime\right)$ is the NDCG loss that results from reversing the order of the points $x$ and $x^\prime$, and the summation is over all the points which relate to the same query. See \cite{quoc2007learning} for more details.

As discussed in Section~\ref{sec:intro} and Section~\ref{sec:overfitting}, the gradient-descent style boosting that MART employs may lead to \overspecialization, and a common approach employed to address this issue is to use shrinkage. Under this method, MART operates as described above when learning the new tree in every iteration. However, before adding this newly learned tree to the ensemble, its leaf values are reduced in magnitude by multiplying them with a constant value in $(0,1)$. Shrinkage helps in alleviating the problem of \overspecialization to a certain extent as we observed in Section~\ref{sec:overfitting}.

We now move on to describing the DART algorithm, which is presented as Algorithm~\ref{alg:DART}. DART diverges from MART at two places. First,  when computing the gradient that the next tree will fit, only a random subset of the existing ensemble is considered. Let us say that the current model $M$ after $n$ iterations is such that $M=\sum_{i=1}^n T_i$, where $T_i$ is the tree learned in the $i$'th iteration. DART first selects a random subset $I\subset \{1,\ldots,n\}$ and creates a model $\hat{M} = \sum_{i\in I} T_i$. Given this model, it learns a regression tree $T$ to predict the inverse derivative of the loss function with respect to this modified model by creating the intermediate dataset $\left\{\left(x, -L^\prime_x\left(\hat{M}(x)\right)\right)\right\}$. 

The second place at which DART diverges from MART is when adding the new tree to the ensemble where DART performs a normalization step. The rationale behind the normalization step is that the new trained tree $T$ is trying to close the gap between $\hat{M}$ and the optimal predictor, however, the dropped trees are also trying to close the same gap. Therefore, introducing both the new tree and the dropped trees will result in the model overshooting the target. Furthermore, assuming that the number of trees dropped from the ensemble to create $I$ that result in the model $\hat{M}$ is $k$, the new tree $T$ has roughly $k$ times larger magnitude than each of the individual trees in the set of dropped trees. Therefore, \algname scales the new tree $T$ by a factor of $\nicefrac{1}{k}$ such that it will have the same order of magnitude as the dropped trees. Following this, the new tree and the dropped trees are scaled by a factor of $\nicefrac{k}{(k+1)}$ and the new tree is added to the ensemble. Scaling by the factor of $\nicefrac{k}{k+1}$ ensures that the combined effect of the dropped trees together with the new tree remains the same as the effect of the dropped trees alone before the introduction of the new tree.

As seen in Figure~\ref{img:contribution} and Figure~\ref{img:trees}, \algname reduces the problem of over-specialization. Therefore, it can be viewed as regularization where the number of trees dropped controls the amount of regularization. On one extreme, if no tree is dropped, DART is no different than MART.  On the other extreme, if all the trees are dropped, the DART is no different than random forest. Therefore, the size of the dropped set allows DART to vary between the ``aggressive" MART mode to a ``conservative" random-forest mode. 

There are many ways to select the trees to be dropped. In the experiments reported here, we have employed what we call the \textit{Binomial-plus-one} technique.  In this technique, each of the existing trees in the ensemble is dropped with a probability \pdrop. However, if no tree was selected to be dropped using the above binomial selection, a single tree is selected uniformly at random to be dropped. Therefore, at least one tree will be dropped at each iteration. 

If \pdrop is set to a very small value, the random selection boils down to simply dropping a single tree in each round. We have experimented with this mode as well, and we denote this mode by defining \pdrop to be $\varepsilon$ in the evaluation results presented in Section~\ref{sec:eval}. 

\begin{algorithm}
\caption{The DART algorithm\label{alg:DART}}
\begin{algorithmic}
\State Let N be the total number of trees to be added to the ensemble
	\State $S_1 \gets \left\{x, -L^\prime_x\left(0\right)\right\}$ 
      \State $T_1$ be a tree trained on the dataset $S_1$ 
	\State $M \gets \{T_1\}$ 
      \For{$t=2,\ldots,N$} 
		\State $D \gets $ the subset of $M$ such that $T\in M$ is in $D$ with probability \pdrop 
		\If {$D=\emptyset$} $D \leftarrow$ a random element from $M$ \EndIf
		\State $\hat{M} \gets M \setminus D$
		\State $S_t \gets \left\{x, -L^\prime_x\left(\hat{M}(x)\right)\right\}$ 
      		\State $T_t$ be a tree trained on the dataset $S_t$
		\State $M \gets M \cup \left\{ \frac{T_t}{|D|+1}\right\}$
		\For{$T \in D$}
			\State Multiply $T$ in $M$ by a factor of $\frac{|D|}{|D| + 1}$
		\EndFor 
	\EndFor
\State Output $M$
\end{algorithmic}
\end{algorithm}

\section{Evaluation} \label{sec:eval}
We evaluated \algname for three different tasks: ranking, regression and classification. For each of the tasks, we used large scale, publicly available datasets. In our evaluation, we compare \algname to MART with different shrinkage factors. Furthermore, since random forests (RF) can be considered as an extreme case of \algname, we compare to this algorithm as well whenever applicable.

\subsection{Ranking}
\begin{table*}
	\centering
	\begin{tabular}{|c|c|c|}
	\hline
       \textbf{Parameter}&\textbf{MART}&\textbf{DART} \\ \hline
       Shrinkage & 0.05, 0.1, 0.2, 0.4 & - \\ \hline
       Dropout rate & - & $\varepsilon$, 0.015, 0.03, 0.045 \\ \hline
       Number of trees & \multicolumn{2}{|c|}{100} \\ \hline
       Leaves per tree & \multicolumn{2}{|c|}{40} \\ \hline
       Loss function parameter  & \multicolumn{2}{|c|}{0.2,0.4,0.6,0.8,1,1.2} \\ \hline
       Fraction of features scanned per leaf  & \multicolumn{2}{|c|}{0.5, 0.75, 1.0} \\ \hline
	\end{tabular}
      \caption{Parameter values scanned for the ranking task. }\label{tab:ranking_param}
\end{table*}
\begin{table*}
	\centering
	\begin{tabular}{|c|c|c|c|c|c|}
		\hline
		\textbf{algorithm} & \textbf{Shrinkage} & \textbf{Dropout} & \textbf{Loss function parameter} & \textbf{Feature fraction} & \textbf{NDCG@3} \\ \hline 
		\textbf{MART} & 0.4 & 0 & 1.2 & 0.75 & 46.31 \\ \hline 
		\textbf{DART} & 1 & 0.03 & 1.2 & 0.5 & \textbf{46.70} \\ \hline

	\end{tabular}
	\caption{NDCG scores for MART and DART on the ranking task. For NDCG scores, higher is better.}\label{tab:ranking_scores}
\end{table*}
MART is commonly used for ranking tasks. For example, in the Yahoo! learning to rank challenge, the winners employed boosted trees \citep{chapelle2011yahoo} based on the LambdaMart method \citep{burges2010ranknet}. We introduced dropouts as explained in Section~\ref{sec:alg} into LambdaMart and tested it on the MSLR-WEB10K dataset.\footnote{\url{http://research.microsoft.com/en-us/projects/mslr/default.aspx}} This dataset contains $\sim1.2M$ query-URL pairs for $10K$ different queries and the task is to rank the URLs for each query according to their relevance using the 136 available features. 

The dataset is partitioned into five parts such that $60\%$ of the data is used for training, $20\%$ is used for validation, and $20\%$ for testing. We scanned the values of various parameters for both algorithms by training on the training data and comparing their performance on the validation data. We selected the best performing models based on their scores on the validation set, and applied them to the test set to obtain the reported results. The different parameters scanned are summarized in Table~\ref{tab:ranking_param}. For each of the parameter combinations experimented, we computed the NDCG score at position 3 and used this as the metric for selecting the parameter values. NDCG \citep{burges2005learning} is a common metric used to evaluate web-ranking tasks. Moreover, the loss functions used were designed to optimize this metric \citep{burges2010ranknet}.

Table~\ref{tab:ranking_scores} presents the main results for the ranking task. DART gains $\sim0.4$ NDCG points over MART.  Moreover, when checking the NDCG scores at positions 1 and 2 we see significant gains as well ($0.2$ points gain in position 1 and $0.38$ points gain in position 2). To put this observed improvement in perspective, in the Yahoo! learning to rank challenge, the gap, in terms of NDCG, between the winners and the team who ranked 5th was 0.35 points~\citep{chapelle2011yahoo}.

\subsection{Regression}\label{sec:regression}
\begin{table*}
	\centering
	\begin{tabular}{|c|c|c|c|}
	\hline
       \textbf{Parameter}&\textbf{MART}&\textbf{\algname}&\textbf{Random Forest} \\ \hline
       Shrinkage & \textbf{0.05}, 0.1, 0.2, 0.3, 0.5 & - & - \\ \hline
       Dropout rate & - & $\varepsilon$, \textbf{0.01}, 0.025, 0.05, 0.1, 0.2& - \\ \hline
       Fraction of instances & \textbf{1.0} & \textbf{1.0} & {0.25, 0.5, 0.75, \textbf{1.0}} \\ 
       used per tree &&&\\ \hline
      Number of trees & \multicolumn{3}{|c|}{25, 50,100,250,500,\textbf{1000}} \\ \hline
       Leaves per tree & \textbf{50},100,250,500,1000 & \textbf{50},100,250,500,1000  & 50,100,250,500,\textbf{1000} \\ \hline
       Fraction of features & 0.05,0.1,\textbf{0.2},0.4, & 0.05,0.1,0.2,\textbf{0.4}, & 0.01,0.025,0.05,0.1,0.2,\textbf{0.4}, \\ 
	scanned per leaf &0.8,1.0&0.8,1.0&0.5,0.8,1.0\\ \hline
 	\end{tabular}
      \caption{Parameter values scanned for the regression task. The parameter values that yielded the lowest loss under each algorithm are highlighted.\label{tab:regression_param}}
\end{table*}

\begin{table*}
	\centering
	\begin{tabular}{|c||c|c|c|c|c|c|}
	\hline
	\textbf{Ensemble size}&25&50&	100&	250&	500&	1000 \\ \hline
	\textbf{MART}&35.13&31.79&30.92&30.07&29.76&29.28\\ \hline
	\textbf{DART}&\textbf{32.50}&\textbf{30.50}&\textbf{29.66}&\textbf{28.14}&\textbf{28.11}&\textbf{27.98}\\ \hline
	\textbf{Random Forest}&32.76&33.21&32.88&32.36&32.66&32.33\\ \hline
	\end{tabular}
	\caption{L2 error of optimal parameter combinations for DART, MART and random forest on the regression task for various ensemble sizes. DART outperforms MART and random forest for all the ensemble sizes tested (the best result for every ensemble size is boldfaced).  \label{tab:regression_results}}
\end{table*}

To test the merits of using dropouts for regression tasks we have used the CT slices dataset \citep{graf20112d} available at the UCI repository \citep{UCI}. This dataset contains 53500 histograms created from CT scans of 74 individuals. The task is to infer the location on the axial axis where the image was taken from. Each image is represented as a 386 dimensional feature vector. We scanned values for various parameters involved and these are summarized in Table~\ref{tab:regression_param}. We have used 10 fold cross validation to compare the algorithms. The folds were selected such that either all the images of an individual are in the train set or all of them are in the test set. 

The evaluation results for the regression task are presented in Table~\ref{tab:regression_results}. For every ensemble size, the best DART model, outperformed both the best MART and the best RF models. We observed that DART outperforms MART and RF even when DART is restricted to drop only a single tree in every iteration (that is the dropout rate is $\varepsilon$).

Furthermore, we observed that RF requires large trees to achieve low losses. For example, when the tree sizes are limited to 50 and 100 leaves, the best RF model achieved a loss of 44.48 and 36.29 respectively. On the other hand, MART and DART achieve their lowest loss values with trees comprising only 50 leaves. 

\subsection{Classification}
\begin{table*}
	\centering
	\begin{tabular}{|c|c|c|c|}
	\hline
       \textbf{Parameter}&\textbf{MART}&\textbf{DART}&\textbf{Random Forest} \\ \hline
       Shrinkage &  0.2, \textbf{0.3}, 0.4, 0.5 & - & - \\ \hline
       Dropout rate & - & $\bm{\varepsilon}$, 0.015, 0.03, 0.045& -\\ \hline
       Fraction of instances  per tree& \textbf{1.0} & \textbf{1.0} & \textbf{0.25}, 0.5, 0.75, 1.0\\ \hline
       Number of trees & 50, 100, \textbf{250}, 500, 1000& 50, 100, \textbf{250}, 500, 1000& 50, 100, 250, \textbf{500}, 1000 \\ \hline
       Leaves per tree & \textbf{40}& \textbf{40}& 50, 100, 250, 500, \textbf{1000} \\ \hline
       Loss function parameter  & 0.2, 0.3, 0.4, \textbf{0.5}& 0.2, 0.3, 0.4, \textbf{0.5} & - \\ \hline
       Fraction of features per leaf  & 0.5, 0.75, \textbf{1.0}& 0.5, 0.75, \textbf{1.0}& 0.5, 0.75, \textbf{1.0} \\ \hline
	\end{tabular}
      \caption{Parameter values scanned for the classification task. The parameter values that yielded the highest accuracy under each algorithm are highlighted. \label{tab:classification_param}}
\end{table*}

\begin{table*}
	\centering
	\begin{tabular}{|c||c|c|c|c|c|}
	\hline
	\textbf{Ensemble size}&	50&	100&	250&	500&	1000 \\ \hline
	\textbf{MART}&	\textbf{0.9687}&	\textbf{0.9699}&	0.9707&\textbf{0.9704}&	0.9695\\ \hline
	\textbf{DART}&	0.9676&	0.9692&	\textbf{0.9714*}&	0.9693&	\textbf{0.9699}\\ \hline
	\textbf{Random Forest} &  0.9627 & 0.9629 & 0.9629& 0.9630 & 0.9628 \\ \hline

	\end{tabular}
	\caption{Accuracies on the test set for DART, MART and random forest on the face-detection classification task for various ensemble sizes. The results are comparable between DART and MART: while MART ``wins" on 3 out of the 5 different ensembles sizes, however, the best model is a DART model.\label{tab:classification_results}}
\end{table*}
The performance of DART on classification tasks was evaluated using the face detection (fd) dataset from the Pascal Large Scale Learning Challenge.\footnote{\url{http://largescale.ml.tu-berlin.de/instructions/}} This dataset contains 30x30 gray scale images and the goal is to infer whether there is a face in the image or not. We used the first 300K examples for training, the next 200K examples for validation and the next 200K examples for testing. The parameters scanned for this task are summarized in Table~\ref{tab:classification_param}. 

We used the validation set to select the best performing parameters for the MART, DART and random forest models and evaluated them on the test set. Table~\ref{tab:classification_results} presents the results for the classification task. Both MART and DART achieve the highest accuracy with ensembles of $250$ trees. Although the difference in accuracies is small, it is statistically significant ($P<0.0001$), since the two models disagree on 1106 predictions on the test set and the MART model gets only 481 of them right while the DART model gets 625 of them correct. The main difference between the models is in their recall where MART has a recall rate of $0.665$ while DART has a recall rate of $0.672$. This is a significant difference for this dataset due to its highly skewed nature: only $\sim8.6\%$ of the instances are labeled positive.  Random forest exhibits lower accuracy for this task. 

In our experiments, random forest did not compare well against MART or DART. Since MART and random forest are the two extremes of the DART algorithm, it serves us to show that the optimal point between these two extremes is not trivial.

\section{Conclusions}
Dropouts~\citep{hinton2012improving} have been shown to improve the accuracies of Neural Network models significantly. On the other hand, Multiple Additive Regression Trees (MART) \citep{friedman2001greedy, JANE:JANE1390} have been found to be the most accurate models for many tasks \citep{caruana2006empirical}, most notably the web ranking task \citep{chapelle2011yahoo}. Motivated by the observation that MART adds trees with significantly diminishing contributions, we hypothesize that dropouts can provide efficient regularization for MART and propose the \algname algorithm. Our experiments show that this is indeed the case: trees in the ensemble created by \algname contribute more evenly towards the final prediction, as shown in Figure~\ref{img:contribution}. In addition, this results in considerable gains in accuracies for ranking, regression and classification tasks. 

This study opens the door to several future directions. For example, using the same technique proposed in this work, it is possible to introduce dropouts in other models such as AdaBoost \citep{freund1995desicion}. The simplicity of these models may allow us to improve our understanding of dropouts. Another direction is to further tune the DART algorithm by experimenting different ways of selecting the dropped set and the normalization techniques. 
Furthermore, the even contribution of the trees in DART may allow using it for learning tasks with drifting targets. This can be achieved, for example, by periodically dropping a subset of the existing trees and learning new trees, with new data, to replace them.

\section*{Acknoledgments}
This research was conducted while the first author was an intern at the machine learning department at Microsoft Research. 
\bibliographystyle{plainnat}
\bibliography{treeDrop}{}

\begin{thebibliography}{18}
\providecommand{\natexlab}[1]{#1}
\providecommand{\url}[1]{\texttt{#1}}
\expandafter\ifx\csname urlstyle\endcsname\relax
  \providecommand{\doi}[1]{doi: #1}\else
  \providecommand{\doi}{doi: \begingroup \urlstyle{rm}\Url}\fi

\bibitem[Bache and Lichman(2013)]{UCI}
Kevin Bache and Moshe Lichman.
\newblock {UCI} machine learning repository, 2013.
\newblock URL \url{http://archive.ics.uci.edu/ml}.

\bibitem[Breiman(2001)]{breiman2001random}
Leo Breiman.
\newblock Random forests.
\newblock \emph{Machine learning}, 45\penalty0 (1):\penalty0 5--32, 2001.

\bibitem[Burges et~al.(2005)Burges, Shaked, Renshaw, Lazier, Deeds, Hamilton,
  and Hullender]{burges2005learning}
Chris Burges, Tal Shaked, Erin Renshaw, Ari Lazier, Matt Deeds, Nicole
  Hamilton, and Greg Hullender.
\newblock Learning to rank using gradient descent.
\newblock In \emph{Proceedings of the 22nd international conference on Machine
  learning}, pages 89--96. ACM, 2005.

\bibitem[Burges(2010)]{burges2010ranknet}
Christopher~JC Burges.
\newblock From ranknet to lambdarank to lambdamart: An overview.
\newblock \emph{Learning}, 11:\penalty0 23--581, 2010.

\bibitem[Burges et~al.(2007)Burges, Ragno, and Viet~Le]{quoc2007learning}
Christopher~J.C. Burges, Robert Ragno, and Quoc Viet~Le.
\newblock Learning to rank with nonsmooth cost functions.
\newblock \emph{NIPS’07}, 19:\penalty0 193, 2007.

\bibitem[Caruana and Niculescu-Mizil(2006)]{caruana2006empirical}
Rich Caruana and Alexandru Niculescu-Mizil.
\newblock An empirical comparison of supervised learning algorithms.
\newblock In \emph{Proceedings of the 23rd international conference on Machine
  learning}, pages 161--168. ACM, 2006.

\bibitem[Chapelle and Chang(2011)]{chapelle2011yahoo}
Olivier Chapelle and Yi~Chang.
\newblock Yahoo! learning to rank challenge overview.
\newblock In \emph{Yahoo! Learning to Rank Challenge}, pages 1--24, 2011.

\bibitem[Elith et~al.(2008)Elith, Leathwick, and Hastie]{JANE:JANE1390}
Jane Elith, John~R Leathwick, and Trevor Hastie.
\newblock A working guide to boosted regression trees.
\newblock \emph{Journal of Animal Ecology}, 77\penalty0 (4):\penalty0 802--813,
  2008.
\newblock ISSN 1365-2656.
\newblock \doi{10.1111/j.1365-2656.2008.01390.x}.
\newblock URL \url{http://dx.doi.org/10.1111/j.1365-2656.2008.01390.x}.

\bibitem[Freund(2001)]{freund2001adaptive}
Yoav Freund.
\newblock An adaptive version of the boost by majority algorithm.
\newblock \emph{Machine learning}, 43\penalty0 (3):\penalty0 293--318, 2001.

\bibitem[Freund and Schapire(1995)]{freund1995desicion}
Yoav Freund and Robert~E Schapire.
\newblock A desicion-theoretic generalization of on-line learning and an
  application to boosting.
\newblock In \emph{Computational learning theory}, pages 23--37. Springer,
  1995.

\bibitem[Friedman(2001)]{friedman2001greedy}
Jerome~H Friedman.
\newblock Greedy function approximation: a gradient boosting machine.
\newblock \emph{Annals of Statistics}, pages 1189--1232, 2001.

\bibitem[Friedman(2002)]{friedman2002stochastic}
Jerome~H Friedman.
\newblock Stochastic gradient boosting.
\newblock \emph{Computational Statistics \& Data Analysis}, 38\penalty0
  (4):\penalty0 367--378, 2002.

\bibitem[Graf et~al.(2011)Graf, Kriegel, Schubert, P{\"o}lsterl, and
  Cavallaro]{graf20112d}
Franz Graf, Hans-Peter Kriegel, Matthias Schubert, Sebastian P{\"o}lsterl, and
  Alexander Cavallaro.
\newblock 2d image registration in ct images using radial image descriptors.
\newblock In \emph{Medical Image Computing and Computer-Assisted
  Intervention--MICCAI 2011}, pages 607--614. Springer, 2011.

\bibitem[Hinton et~al.(2012)Hinton, Srivastava, Krizhevsky, Sutskever, and
  Salakhutdinov]{hinton2012improving}
Geoffrey~E Hinton, Nitish Srivastava, Alex Krizhevsky, Ilya Sutskever, and
  Ruslan~R Salakhutdinov.
\newblock Improving neural networks by preventing co-adaptation of feature
  detectors.
\newblock \emph{arXiv preprint arXiv:1207.0580}, 2012.

\bibitem[Krizhevsky et~al.(2012)Krizhevsky, Sutskever, and
  Hinton]{krizhevsky2012imagenet}
Alex Krizhevsky, Ilya Sutskever, and Geoffrey~E Hinton.
\newblock Imagenet classification with deep convolutional neural networks.
\newblock In \emph{Advances in neural information processing systems}, pages
  1097--1105, 2012.

\bibitem[Maaten et~al.(2013)Maaten, Chen, Tyree, and
  Weinberger]{ICML2013_vandermaaten13}
Laurens Maaten, Minmin Chen, Stephen Tyree, and Kilian~Q. Weinberger.
\newblock Learning with marginalized corrupted features.
\newblock In Sanjoy Dasgupta and David Mcallester, editors, \emph{Proceedings
  of the 30th International Conference on Machine Learning (ICML-13)},
  volume~28, pages 410--418. JMLR Workshop and Conference Proceedings, 2013.
\newblock URL
  \url{http://jmlr.csail.mit.edu/proceedings/papers/v28/vandermaaten13.pdf}.

\bibitem[Wager et~al.(2013)Wager, Wang, and Liang]{wager2013dropout}
Stefan Wager, Sida Wang, and Percy Liang.
\newblock Dropout training as adaptive regularization.
\newblock In \emph{Advances in Neural Information Processing Systems}, pages
  351--359, 2013.

\bibitem[Wang and Manning(2013)]{wang2013fast}
Sida Wang and Christopher Manning.
\newblock Fast dropout training.
\newblock In \emph{Proceedings of the 30th International Conference on Machine
  Learning (ICML-13)}, pages 118--126, 2013.

\end{thebibliography}
\end{document}